\documentclass[10pt,twocolumn,letterpaper]{article}

\usepackage{cvpr}
\usepackage{times}
\usepackage{epsfig}
\usepackage{graphicx}
\usepackage{amsmath}
\usepackage{amssymb}

\usepackage[breaklinks=true,bookmarks=false]{hyperref}

\cvprfinalcopy 

\def\cvprPaperID{****} 

\begin{document}

\title{Effective Approaches to Batch Parallelization for \\ Dynamic Neural Network Architectures}

\author{Joseph Suarez*\\
{\tt\small joseph15@stanford.edu}
\and
Clare Zhu*\\
{\tt\small clarezhu@stanford.edu}
}

\maketitle

\begin{abstract}

We present a simple dynamic batching approach applicable to a large class of dynamic architectures that consistently yields speedups of over 10x. We provide performance bounds when the architecture is not known a priori and a stronger bound in the special case where the architecture is a predetermined balanced tree. We evaluate our approach on Johnson et al.'s recent visual question answering (VQA) result of his CLEVR dataset by Inferring and Executing Programs (IEP). We also evaluate on sparsely gated mixture of experts layers and achieve speedups of up to 1000x over the naive implementation.

\end{abstract}

\section{Introduction}
\makeatletter{\renewcommand*{\@makefnmark}{}
\footnotetext{* Authors contributed equally}\makeatother}
The problem we address is neither VQA nor optimization of a single architecture. Our motivation is to accelerate a large class of dynamic architectures such that they become computationally comparable to their static counterparts. This cause is not motivated only by the recent successes of dynamic architectures, but by their numerous desirable properties that make them likely to retain and increase in importance in the future, particularly their ability to explicitly modularize knowledge.

We specifically explore Johnson et al.'s recent work\cite{DBLP:journals/corr/JohnsonHMHLZG17} in greatest detail because it serves as a useful testbed for multiple approaches to dynamic batching. Their execution engine's modules (see \hyperref[sec:background]{Background} or original work) not only yield dramatic accuracy gains over all strong baselines, but are also a prime example of explicit modularization of knowledge. We view this as a key advantage of dynamic architectures directly comparable to facets of human intelligence. Our work successfully enables efficient parallelization over minibatches in a large class of architectures despite the fact that a new network is assembled for each example.  

\subsection{Related Work}

Previous notable dynamic graph results include neural module networks \cite{Andreas_2016_CVPR}, which form the basis of the execution engine of Johnson et al. in their CLEVR\cite{DBLP:journals/corr/JohnsonHMFZG16} IEP result. The difference is that latter's architecture is built on generic, minimally-engineered neural network blocks that are more likely to generalize to a wider class of problems than the original neural module networks approach, which uses a heavily-engineered question parser and custom per-module architectures. Whereas improvement upon neural module networks constitutes improvement upon a single architecture, improvement on the CLEVR architecture is generalizable to a wide class of models under a minimal set of assumptions (see \hyperref[sec:discussion]{Discussion}).

Additional dynamic graph results include neural Turing machines \cite{graves2014neural} \cite{graves2016hybrid} and memory networks \cite{weston2014memory} \cite{NIPS2015_5846},  which both provide auxiliary queryable memory for read/write use during inference. While such architectures are applicable in problems requiring long-term memory, visual question answering places more focus on short term memory. Like the IEP result, these works tend towards higher level reasoning. However, they are perhaps less directly comparable than approaches that explicitly attempt to build generalizable program structures, such as neural program interpreters \cite{reed2015neural}. The main difference is that the IEP result assembles programs that are defined in their entirety before being executed, thus additional dynamic batching optimizations are possible. Note that a subset of our results are applicable in both cases.

\subsection{Background}
\label{sec:background}

Much of our work is built atop the recently published CLEVR dataset and subsequent IEP result. We briefly outline these for convenience.

\subsubsection{CLEVR}

CLEVR is a VQA dataset comprising 70K images and 700K questions/answers/programs triplets. Images are synthetic but high quality 3D renders of geometric objects with varying shapes, sizes, colors, and textures. The standard VQA task is given by (question, image) $\to$ (answer). The difference lies in the inclusion of \textit{programs} in CLEVR, which are functional representations of the questions. CLEVR therefore allows VQA to be split between two intermediate tasks, as in the IEP result: (question) $\to$ (program) and (program, image) $\to$ (answer).

One might argue that intermediate programs are unrealistic, as one is unlikely to have program annotations in large, realistic tasks. From the CLEVR result, it seems likely that one could collect a small number of annotations on realistic datasets and use these to initialize the program generator. This is similar to the transfer learning experiment in the IEP result. However, performance did degrade compared to the original task; additional work is required to close the gap.

\subsubsection{Visual Reasoning Programs}

The IEP result consists of a program generator and execution engine. The program generator is a 2-layer word-level question encoder LSTM \cite{doi:10.1162/neco.1997.9.8.1735} and 2-layer word (function)-level program decoder LSTM. We focus on the execution engine, as it is the dynamic portion of the architecture and the source of the majority of computation time.

The program generator predicts a sequence of functions over the function vocabulary with a standard argmax. As the arity (number of arguments) of each function is predetermined, there exists a unique mapping from the predicted vector of functions to a program tree. This is assembled via a depth-first search. Each function is itself a neural network, with the exception of a special \textit{SCENE} token, which instead outputs ResNet-101 features \cite{He_2016_CVPR} taken from an intermediate layer. This program tree is then directly executed, and the outputs are passed through a small classifier network (one convolutional and two fully connected layers) to yield a softmax confidence distribution over answers, which is then optimized as normal via backpropagation over the cross-entropy loss.

\section{Methods}
 
In the IEP result, programs must be executed sequentially with an explicit loop over the examples in each minibatch. As a result, unlike static networks, the computation time of the forward pass scales linearly with the batch size. We present two variants of topological sort that remedy this issue.

To clarify the ongoing notation, programs have max length $s$ and function vocabulary size $p$. The batch size is denoted by $b$ and the max program tree depth by $d$. 

\textbf{Standard topological sort.} First, consider a naive topological sort. Each program tree is sorted via an infix depth-first search. This results in a queue ordering such that each node can be executed sequentially; no node is executed before all of its dependents. While this operation runs in time linear in the number of nodes (e.g. $O(bs)$), it is fast compared to expensive neural network operations and can be multithreaded extremely efficiently, thus we ignore this factor in our computations.

We now have a flat representation of each program, which can be viewed as a grid of size $b \times s$. Instead of executing each program independently, we loop only over the rows and execute one full column of size $b$. Each node corresponds to a different element in the function vocabulary. However, for $b>p$, we need only make at most $p$ expensive neural network calls instead of $b$. This results in $O(ps)$ neural network calls.

\textbf{Improved topological sort.} In the improved variant of topological sort, we take this sorting operation one step further. Instead of flattening programs, we instead label each node by its maximum distance from the root node. Nodes with the same label are pooled. Each pool is executed at once in $O(p)$ neural network calls, for a total of $O(pd)$ calls. In the case where program trees are balanced (important in the design of future datasets), this yields $O(p \log_2 s)$ execution. The program trees used in CLEVR are, unfortunately, highly imbalanced, thus this approach results in only a 10-25 percent speedup over standard topological sort. Note that, as $d$ is the \textit{maximum} depth across all programs in a minibatch, $d \gg \log_2 s$. 

\section{Results}

\begin{figure}
	\centering
	\includegraphics[scale=0.325]{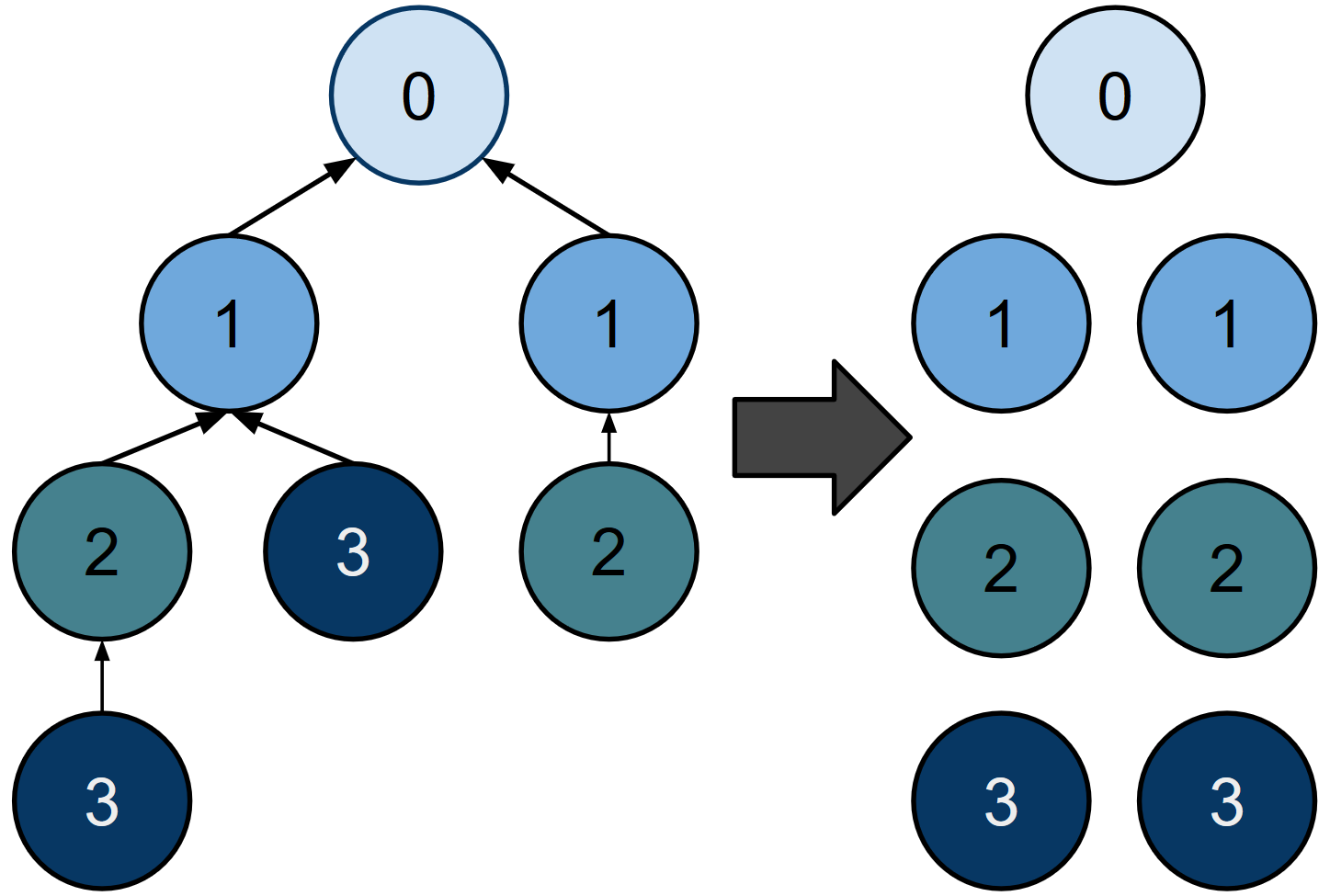}
	\caption[figure]{Example of program tree labeling scheme. Nodes are labeled and aggregated based upon their dependencies.}
\end{figure}

\subsection{IEP}

We evaluate performance gains with improved topological sort vs. our implementation of the original IEP architecture. Relevant portions of program construction/execution code are shared appropriately: our experiments are robust to any unintended inefficiency in our implementation of the original architecture.

Using all memory in a single Nvidia GTX 1080Ti, we achieve 5.5X faster inference and a 2X faster backward pass. It is currently unclear why gains do not better transfer to the backwards pass in the PyTorch backend, as the expected gains are symmetric. However, this is not a fair comparison, as over half of computation time is spent in inefficient CPU Python graph sorting code. Furthermore, this CPU code is embarrassingly parallel and should be written in multithreaded C++. 

Thus, a fairer comparison is to measure neural cell execution time, in which case we achieve over 14X gains (see Fig. 2). There is a small amount of additional data stacking code omitted from this computation because it can likely also be optimized and is not directly comparable to the original architecture; unfairly including this, gains are still well over 10X.

More importantly, scaling is linear with batch size. Doubling GPU memory yields 2X performance. Those familiar with minibatch parallelism may object that this performance gain usually drops off after a certain point (minibatch size $>$1000 in our experience). However, this is not likely to be an issue in our case, as there is an additional factor of the program function vocabulary size (40). With equal distribution of execution over functions, minibatch size 1000 per cell corresponds to overall minibatch of size 40000, which would require approximately 500 GB of GRAM. While increasing the program vocabulary does incur a linear decrease in performance, it causes an equivalent increase in maximum minibatch size before incurring diminishing returns.

Furthermore, it is possible to maintain such gains in the case of multiple GPUs (e.g. large batch size split over many devices) by assigning a different cell function to each GPU. Goyal et. al recently demonstrated that this sort of data parallel scaling can remain practical even at extremely large batch sizes by scaling the learning rate correspondingly \cite{goyal2017accurate}.

\begin{figure}
	\centering
	\includegraphics[scale=0.29]{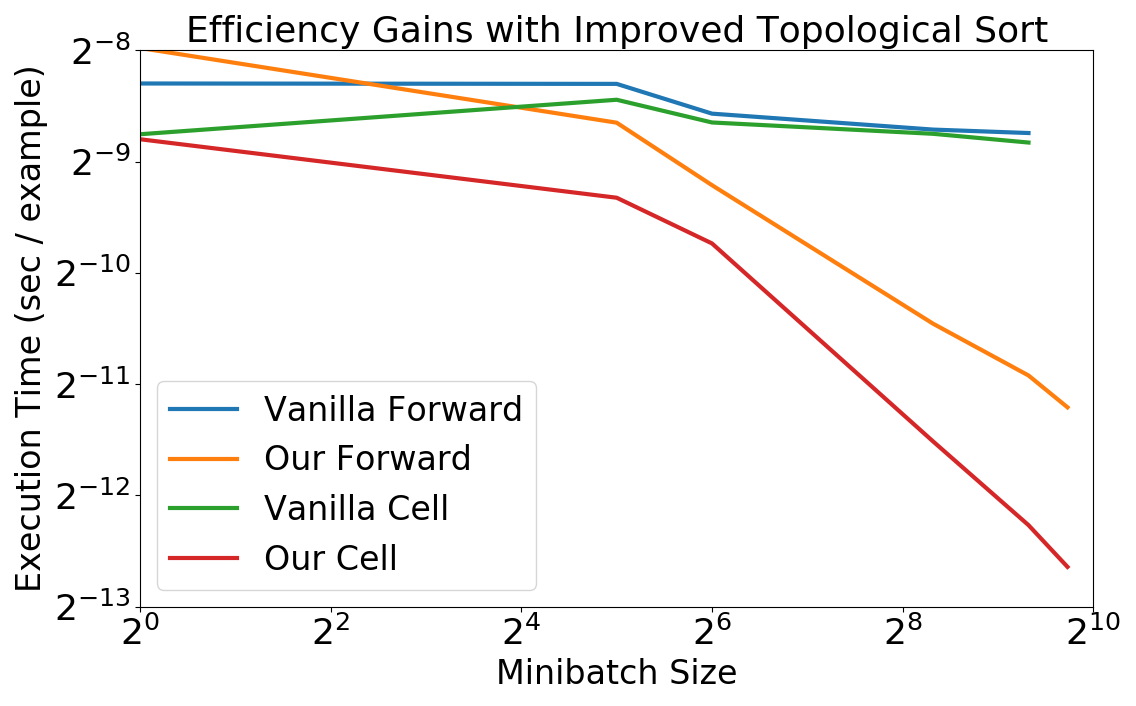}
	\caption[figure]{Log scaled visualization of efficiency gains incurred from our improved topological sort. Vanilla denotes the implementation in Johnson et al. This makes clearer the near-linear gains in speed as the minibatch size approaches 1000.}
\end{figure}

\subsection{Sparsely Gated Mixture of Experts}

As a second test, we evaluate performance on the sparsely gated mixture of experts (MOE) layer as in \cite{shazeer2017outrageously}, where we maintain a set of \textit{n} fully connected expert networks of which \textit{k} are active for each of the \textit{b} examples in each minibatch, and $k \ll n$. Note that unlike the IEP example above, each MOE layer must fully terminate before the next layer can begin execution: the architecture is not known a-priori. We therefore apply a degenerate case of our standard topological sort: we batch computation across all currently known modules. For the MOE layer, this corresponds to $k$ experts in each of $b$ examples for a total of $kb$ known modules.

The naive implementation loops over and executes the $kb$ experts independently. Our implementation makes only $n$ expert calls. We evaluate vs. the naive implementation using 256-dimensional data. Each expert is a neural network with one hidden layer. We tested many network sizes, but found that network size is largely independent of the speedup factor. We therefore only present results with 256 hidden units in Fig. 3; this is the largest experiment that fit on a single 11GB card.

Our metric for performance is cell execution speed (e.g. total time spent executing experts) rather than forward pass speed. This is, as in IEP, to avoid unfair comparison to large swaths of unoptimized python CPU code. Note that there is an additional data stacking operation in our approach omitted from these calculations, as in the IEP result, because including it is not an equal comparison to the vanilla approach. Also, this stacking operation is almost certainly required for any approach in the distributed case.

Notice that performance quickly decays as the number of experts approaches the minibatch size. This should be expected, as parallelization is impossible if each expert receives one example. It is difficult to predict the largest practical speedup achievable, as we do not possess the computational resources to run production-scale results. However, we can extrapolate from the results above. Furthermore, we assume that our results will scale linearly with many GPUs. This is reasonable both from our suggested parallelization scheme (see IEP) and the results of the similar scheme utilized in the original MOE work.

Consider the case where each of \textit{n} experts has \textit{h} hidden units, and our data is \textit{d} dimensional. This yields $2hnd$ parameters. If we want each expert to to receive \textit{m} examples and we use \textit{k} experts per example, then we must store $\frac{nm(2d + h)}{k}$ floating point activations. Thus the ratio of memory required between the activations and the parameters is $\frac{m(2d+h)}{2khd}$ Note that this does not depend on the number of experts.

As a practical example, with $h=d=2048$, $k=100$, and $m=1$ million, this ratio is 7.3. Thus even outrageously large networks require a reasonable fraction of the total memory in order to use extremely large batch sizes. With 10k experts, we would expect 50-80X performance with this configuration. Furthermore, we could increase $k$ by up to a factor of 10 if we desired without significant loss of absolute performance and obtain a 8-10X efficiency gain.

\begin{figure}
	\centering
	\includegraphics[scale=0.35]{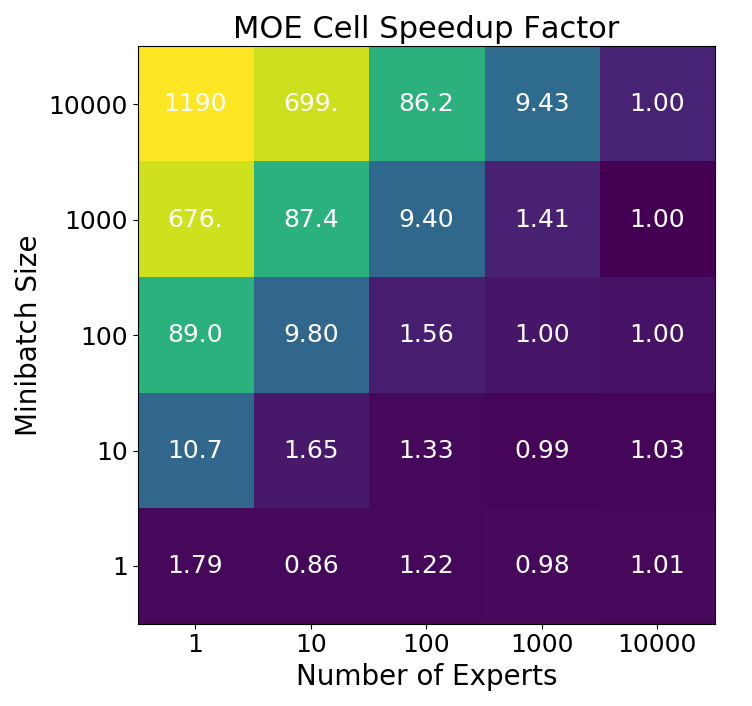}
	\caption[figure]{Cell execution speedup time with dynamic batching in a sparsely gated mixture of experts layer. Speedup factor is computed by averaging over only 5 runs for efficiency. Each expert is a 256-256-\#experts fully connected network.}
\end{figure}

\section{Discussion}
\label{sec:discussion}

The recent independent results of automatic batching in DyNet \cite{neubig2017fly} and the TensorFlow Fold library \cite{looks2017deep} are  closest to our work. However, the DyNet batching optimizer relies on lazy execution to optimally organize data on the fly, and TensorFlow Fold similarly operates over compiled graphs. 

Lazy execution is not present in alternative frameworks such as PyTorch by design, as it is decouples implementation from execution, which is often an undesirable quality during implementation. Without lazy execution, automatic batching is not possible, but variants of our approach are still viable. Our standard and improved topological sorting approaches can be viewed as a set of manual batching optimizations applicable to a large class of dynamic architectures without requiring lazy execution or prior compilation.

With regard to technical implementation, we mark the dependencies of each cell as in DyNet. However, we batch by dependency depth as in TensorFlow Fold in order to minimize extra CPU code. This facilitates changing between approaches. While our experiments are not directly comparable to the benchmarks vs. TensorFlow Fold in the DyNet result, our speedup curves maintain the linear gains of TensorFlow Fold at large batch sizes.

Our improved topological sort makes the following hard assumptions in order to achieve $O(p \log_2 s)$ where complexity is measured by the number of calls to expensive (e.g. neural network) functions:
\begin{enumerate}
	\item There exists a set of $p$ expensive modules.
    \item The architecture, composed of modules, can be executed with batch size $b$ such that $b \gg p$.
    \item The architecture is a balanced tree with structure and module arities known a priori.
\end{enumerate}

In the case where the third assumption fails because the architecture (and arities) are known ahead of time but it is a general DAG, our improved topological sort is still applicable, but with complexity $O(pd)$ where $d$ is the maximum dependency path length. This has the same complexity as the standard topological sort but with an equivalent or more favorable constant factor which can be fairly significant, depending on the number of concurrent branches in the graph.

In the case where the architecture is a generic graph but is not specifically known ahead of time, it is no longer possible to use our improved topological sort. However, as the next module is always known in any architecture, it is still possible to apply our standard topological sort approach and achieve $O(pd)$ by aggregating the computations of all current modules over $p$, as done in the MOE example above. In the presence of cycles, $d$ becomes the maximum length of an unrolled graph. This is always limited in practice to avoid infinite cycles.

In general, our approach is applicable whenever significant module reuse is present among examples. While it may be possible to improve upon our approach in select cases by searching over the set of known modules to optimize the order of dependency execution, this would require additional CPU code that may not be possible to fully optimize.

\section{Conclusion}

We demonstrate the effectiveness of our dynamic batching method on IEP and MOE (our codebase is available at \url{https://github.com/jsuarez5341/Efficient-Dynamic-Batching}), achieving over 14X and up to 1000X neural cell execution, respectively. In each case, we characterize the trend of improvements as batch size varies, which yields increasing returns and becomes linear until extremely large batch size. We define the class of problems for which our improved topological sort is applicable as well as the class where it is not but standard topological sort is still feasible; in both cases, we provide complexity bounds as a function of neural network calls. The breadth of architectures in which at least one variant of our approach is applicable implies that a large class of dynamic architectures can be trained and executed as quickly and efficiently as their static counterparts.

{\small
\bibliographystyle{ieee}
\bibliography{egbib}
}

\end{document}